\documentclass[11pt,a4paper]{article}
\usepackage{authblk} 
\usepackage[hyperref]{acl2019}

\usepackage{times}
\usepackage{latexsym}
\usepackage{tabularx}
\usepackage{graphicx}
\usepackage{csquotes}
\usepackage{booktabs}
\usepackage{multicol}
\usepackage{url}
\usepackage{xspace}

\aclfinalcopy 

\newcommand\corpusname{CompSent-19\xspace}
\newcommand{\Ni}{(1)~}
\newcommand{\Nii}{(2)~}
\newcommand{\class}[1]{{\texttt{\small #1}}\xspace}

\title{Categorizing Comparative Sentences}

\author[$\star,\ddag$]{\textbf{Alexander Panchenko}}
\author[$\dag$]{\textbf{Alexander Bondarenko}}
\author[$\ddag$]{\textbf{Mirco Franzek}}
\author[$\dag$]{\\\textbf{Matthias Hagen}}
\author[$\ddag$]{\textbf{Chris Biemann}}

\affil[$\star$]{Skolkovo Institute of Science and Technology, Moscow, Russia}
\affil[$\ddag$]{Language Technology Group, Universit{\"a}t Hamburg, Hamburg, Germany}
\affil[$\dag$]{Big Data Analytics Group, Martin-Luther Universit{\"a}t Halle-Wittenberg, Halle, Germany}

\date{}

\begin{document}
\maketitle

\begin{abstract}
We tackle the tasks of automatically identifying comparative sentences and categorizing the intended preference (e.g., ``Python has better NLP libraries than MATLAB'' $\rightarrow$ Python, better, MATLAB). To this end, we manually annotate 7,199~sentences for 217~distinct target item pairs from several domains (27\%~of the sentences contain an oriented comparison in the sense of ``better'' or ``worse''). A gradient boosting model based on pre-trained sentence embeddings reaches an F1 score of~85\% in our experimental evaluation. The model can be used to extract comparative sentences for pro/con argumentation in comparative / argument search engines or debating technologies.   
\end{abstract}

\section{Introduction}

Everyone faces choice problems on a daily basis: from choosing between products (e.g., which camera to buy), to more generic preferences for all kinds of things: cities to visit, universities to study at, or even programming languages to use. Informed choices need to be based on a comparison and objective argumentation to favor one of the candidates. Often, people seek support from other people---for instance, a lot of questions like ``How does X compare to Y?'' are asked on question answering platforms. 

The Web also contains pages about comparing various objects: Specialized web resources systematize human experts results for domain-specific comparisons (for insurances, cameras, restaurants, hotels, etc.) while systems like WolframAlpha aim at providing comparative functionality across domains. Still, such pages and systems usually suffer from coverage issues relying on structured databases as the only source of information ignoring the rich textual content available on the web.

No system is currently able to satisfy open-domain comparative information needs with sufficient coverage and explanations of the compared items' relative qualities. Indeed, information retrieval systems and web search engines are able to directly answer many factoid questions (one-boxes, direct answers, etc.) but do not yet treat comparative information needs any different than standard queries. Search engines show the default ``ten blue links'' for many comparative information needs even though a direct answer enriched by pro/cons for the different options might be the much more helpful result.

One reason might be that despite the wealth of comparisons on the web with argumentative explanations, there is still no widespread technology for its extraction. In this work, we propose the first steps towards closing this gap by proposing classifiers to identify and to categorize comparative sentences. 

The task of identifying and categorizing comparative sentences is to decide for a given sentence whether it compares at least two items and, if so, which item ``wins'' the comparison. For instance, given the sentence \emph{Python is better suited for data analysis than MATLAB due to the many available deep learning libraries}, the system should categorize it as comparative and that it favors Python (Python ``wins'' over MATLAB). Identifying and categorizing comparative sentences can be viewed as a sub-task of argumentation mining~\cite{Lippi2016Argumentation-M} in the sense that detected comparative sentences (and probably also their context sentences) can support pro/con analyses for two or more items. 
Such comparative pro/cons might be used to trigger reactions in debates (one advantage of some item can be countered by some advantage of the other item, etc.) or they can form the basis for answering comparative information needs submitted to argument search engines.

Our main contributions are two-fold: 

\begin{enumerate}
\item We release \corpusname, a new corpus consisting of 7,199~sentences containing item pairs (27\%~of the sentences are tagged as comparative and annotated with a preference);

\item We present an experimental study of supervised classifiers and a strong rule-based baseline from prior work.
\end{enumerate}

The new \corpusname corpus,\footnote{\href{https://zenodo.org/record/3237552}{zenodo.org/record/3237552}} pre-trained sentence categorization models, and our source codes\footnote{\href{https://github.com/uhh-lt/comparative}{github.com/uhh-lt/comparative}} are publicly available online.

\section{Related Work}
\label{sec:relwork}

A number of online comparison portals like GoCompare or Compare.com provide access to structured databases where products of the same class can be ranked along with their aspects. Other systems like Diffen.com and Versus.com try to compare any pair of items on arbitrary properties. They reach high coverage through the integration of a large number of structured resources such as databases and semi-structured resources like Wikipedia, but still list aspects side by side without providing further verbal explanations---none of the portals aim at extracting comparisons from text. Promising data sources for textual comparisons are question answering portals like Quora or Yahoo!\,Answers that contain a lot of ``How does X compare to Y?''-questions with human answers but the web itself is an even larger source of textual comparisons. 

Mining and categorizing comparative sentences from the web could support search engines in answering comparative queries (with potential argumentation justifying the preference in the mined sentence itself or in its context) but also has opinion mining \cite{ganapathibhotla-liu2008} as another important application. Still, previous work on recognizing comparative sentences has mostly been conducted in the biomedical domain. For instance, \citet{fiszman2007interpreting} identify sentences explicitly comparing elements of drug therapy via manually developed comparative and direction patterns informed by a lot of domain knowledge. Later, \citet{park2012identifying} trained a high-precision Bayesian Network classifier for toxicology publications that used lexical clues (comparatives and domain-specific vocabulary) but also paths between comparison targets in dependency parses. More recently, \citet{gupta2017identifying} described a system for the biomedical domain that also combines manually collected patterns for lexical matches and dependency parses in order to identify comparison targets and comparison type using the  as gradable, non-gradable, superlative-taxonomy of \citet{JindalLiu2006}.

Developing a system for mining comparative sentences (with potential argumentation support for a preference) from the web might utilize specialized jargon like hashtags for argumentative tweets \cite{Dusmanu2017Argument-Mining} but at the same time faces the challenges recognized for general web argument mining \cite{Snajder2017Social-Media-Ar}: web text is typically not well formulated, misses argument structures, and contains poorly formulated claims. In contrast to the use of dependency parses for mining comparative sentences in the biomedical domain, such syntactic features are often impossible to derive for noisy web text and were even shown to not really help in identifying argument structures from well-formulated texts like persuasive essays or Wikipedia articles \cite{Aker2017What-works-and-,Stab2014Identifying-Arg};  simpler structural features such as punctuation subsumed syntactic features in the above studies.

The role of discourse markers in the identification of claims and premises was discussed by \citet{Eckle-Kohler2015On-the-Role-of-}, who found such markers to be moderately useful for identifying argumentative sentences. Also \citet{Daxenberger2017What-is-the-Ess} noted that claims share lexical clues across different datasets. They also concluded from their experiments that typical argumentation mining datasets were too small to unleash the power of recent DNN-based classifiers; methods based on feature engineering still worked best. 

\section{Dataset}
\label{sec:dataset}

As there is no large publicly available cross-domain dataset for comparative argument mining, we create one composed of sentences annotated with markers \class{BETTER} (the first item is better or ``wins'') / \class{WORSE} (the first item is worse or ``looses'') or \class{NONE} (the sentence does not contain a comparison of the target items). The \class{BETTER}-sentences represent a pro argument in favor of the first compared item (or a con argument for the second item) while the roles are exchanged for the \class{WORSE}-sentences. 

In our dataset, we aim to minimize domain-specific biases to rather capture the nature of comparison and not the nature of particular domains. We thus decided to control the specificity of domains via the selection of the comparison targets. We hypothesized and could confirm in preliminary experiments that comparison targets usually have a common hypernym (i.e., they are instances of the same class), which we utilize for the selection of the compared item pairs.

The most specific domain we choose is \emph{Computer Science} with comparison targets like programming languages, database products and technology standards such as Bluetooth or Ethernet. Many computer science concepts can be compared objectively (e.g., via transmission speed or suitability for certain applications). The comparison targets were manually extracted from Wikipedia ``List of''-articles that cover computer science. In the annotation process, annotators were asked to label sentences from this domain only if they had some basic knowledge in computer science. 

The second, broader domain is \emph{Brands}. It contains items of various types (e.g., cars, electronics, or food). As brands are present in everyday life, we assume basically anyone to be able to label sentences containing well-known brands such as Coca-Cola or Mercedes. Again, target items for this domain were manually extracted from Wikipedia ``List of''-articles.

The third \emph{Random} domain is not restricted to any topic. For each of 24~randomly selected seed words,\footnote{Created using \href{https://www.randomlists.com}{randomlists.com}: book, car, carpenter, cellphone, Christmas, coffee, cork, Florida, hamster, hiking, Hoover, Metallica, NBC, Netflix, ninja, pencil, salad, soccer, Starbucks, sword, Tolkien, wine, wood, XBox, Yale.} 10~similar words were collected based on the distributional similarity JoBimText API \cite{BR2D13}.

Especially for brands and computer science, the resulting item lists were large (4,493~in brands and 1,339~in computer science). In a manual inspection, low-frequency and ambiguous items were removed (e.g., the computer science concepts \enquote{RAID} (a hardware concept) and \enquote{Unity} (a game engine) are also regularly used nouns). The remaining items were combined into pairs. For each item type (seed Wikipedia list or seed word), all possible item combinations were created. These pairs were then used to mine sentences containing both items from a web-scale corpus.

Our sentence source is the publicly available index of the DepCC \cite{Panchenko:2017aa}, an index of more then 14~billion dependency-parsed English sentences from the Common Crawl filtered for duplicates. This index was queried for sentences containing both items in each target pair. For 90\%~of the pairs, we also added frequent comparative cue words\footnote{Better, easier, faster, nicer, wiser, cooler, decent, safer, superior, solid, terrific, worse, harder, slower, poorly, uglier, poorer, lousy, nastier, inferior, mediocre.} to the query in order to bias the results towards actual comparative sentences but at the same time also allow for comparisons that do not contain any of the anticipated cues. This focused querying was necessary as a random sampling would have resulted in only a very tiny fraction of comparative sentences. Note that even sentences containing a cue word do not necessarily express a comparison between the desired targets (e.g., dog vs.\ cat: He's the best pet that you can get, \textit{better} than a dog or cat). It is thus especially crucial to enable a classifier to learn not to rely on the presence of the cue words only (which is very likely in a random sample of sentences with very few comparisons). For our dataset, we keep target pairs with at least 100~retrieved sentences.

From all sentences for the target pairs, we randomly sampled 2,500~instances in each category as potential candidates for a crowd-sourced annotation that we conducted on the Figure Eight platform in several small batches. Each sentence was annotated by at least five trusted workers. Of all annotated sentences, 71\%~received unanimous votes, and at least~4 out of~5 workers agreed for over~85\%, at least~4 out of~5 workers agreed.

\begin{table*}[t]
\caption{Examples sentences for the three domains with their annotated comparative label (the \textbf{{\color[HTML]{9A14B2}first item}} is \class{BETTER/WORSE/NONE} than the \textbf{{\color[HTML]{6CB219}second item}} (note that the item order matters).}
\label{table-examples}
\centering
\begin{tabularx}{\textwidth}{@{}lXl@{}}
\toprule
\bf Domain & \bf Sentence & \bf Label \\ \midrule

CompSci & This time \textbf{{\color[HTML]{9A14B2}Windows 8}} was roughly 8 percent slower than \textbf{{\color[HTML]{6CB219}Windows 7}}. & \class{WORSE}  \\ 

CompSci & I've concluded that it is better to use \textbf{{\color[HTML]{9A14B2}Python}} for scripting rather than \textbf{{\color[HTML]{6CB219}Bash}}. & \class{BETTER}  \\

Brands & These include \textbf{{\color[HTML]{9A14B2}Motorola}}, Samsung and \textbf{{\color[HTML]{6CB219}Nokia}}. & \class{NONE} \\  

Brands & \textbf{{\color[HTML]{9A14B2}Honda}} quality has gone downhill, Hyundai or \textbf{{\color[HTML]{6CB219}Ford}} is a much better value. & \class{WORSE} \\ 

Random & Right now, I think \textbf{{\color[HTML]{9A14B2}tennis}} is easier than \textbf{{\color[HTML]{6CB219}baseball}}. & \class{BETTER} \\

Random & I've grown older and wiser and avoid the \textbf{{\color[HTML]{9A14B2}pasta}} and \textbf{{\color[HTML]{6CB219}bread}} like the plague. & \class{NONE} \\

\bottomrule                              
\end{tabularx}

\end{table*}

\begin{table}[t]
\caption{Characteristics of our \corpusname dataset.}
\label{table-dataset-characteristics}
\centering
\scalebox{0.95}{
\small
\begin{tabular}{@{}p{2cm}rrr@{\hspace{2em}}r@{}}
\toprule
 & \multicolumn{3}{c}{\bf Label} & \\
 
{\bf Domain} & \class{BETTER} & \class{WORSE} & \class{NONE} & Total \\
\midrule
CompSci & 581 & 248 & 1,596 & 2,425 \\
Brands & 404 & 167 & 1,764 & 2,335 \\
Random & 379 & 178 & 1,882 & 2,439 \\
\midrule 
Total &  1,364 & 593 & 5,242 & 7,199 \\
\bottomrule
\end{tabular}
}
\end{table}

Our final Comparative Sentences Corpus 2019 (\corpusname) is formed by the 7,199~sentences for 271~distinct item pairs that remained after removing the 301~sentences with an annotation confidence below~50\%, a Figure-Eight-internal measure combining annotator trust and voting. Table~\ref{table-examples} shows example sentences with their annotation while Table~\ref{table-dataset-characteristics} outlines the corpus characteristics. Only a 27\%-minority of the sentences are annotated as comparative (despite the selection bias with comparative cue words); in 70\%~of these, the favored item is named first.

\section{Supervised Categorization of Comparative Sentences}
\label{sec:exp}

We split the 7,199~sentences of our \corpusname corpus into an 80\%~training set (5,759 sentences: 4,194 \class{NONE}, 1,091 \class{BETTER}, and 474 \class{WORSE}) and a 20\%~held-out set. During development, the experiments were evaluated on the training set using stratified $5$-fold cross-validation; the held-out set was only used for the final evaluation. If not stated otherwise, scikit-learn \cite{scikit-learn} was used to perform feature processing, classification, and evaluation.

\subsection{Preprocessing}

A first preprocessing step decides if the full sentence or only a part of it should be used for feature computation. Each sentence is considered to consist of three parts: the \textit{beginning part} are all words before the first comparison target, the \textit{ending part} are all words after the second comparison target, and the \textit{middle part} are all words between the targets. Different combinations of partial sentence representations were used in our classification experiments.

The second preprocessing step is carried out to examine the importance of the lexicalized comparison targets for the classification. The targets either stay untouched, are removed, or replaced using two different replacement strategies. In the first variant, both targets are replaced by the term ITEM (\textit{oblivious replacement}). In the second variant, the first object was replaced by ITEM\_A and the second by ITEM\_B (\textit{distinct replacement}). 


\subsection{Supervised Classification Models}

We compare 13~models ranging from the lower-capacity linear models, such as Logistic Regression, Na{\"i}ve Bayes, and SVMs with various kernels to high-capacity ones based on decision trees and their ensembles such as Random Forest, Extra Trees, and Gradient Boosting relying on decision trees. Implementation-wise, twelve of the tested models are available via scikit-learn, while for XGBoost we used the implementation of \citet{DBLP:journals/corr/ChenG16}. Apart from XGBoost and the Extra Trees Classifier, all models have been used in previous argumentation mining studies.

\subsection{Sentence Representations}
\label{sec:features}
We study the classification performance impact of various feature types.

\paragraph{Bag of Words and Bag of Ngrams}
The bag-of-words (BOW) model is a simple vector representation of text documents. All distinct words from the corpus form the vocabulary~$V$. Typically, a document~$d$ is represented by a $V$-dimensional vector~$\mathbf{d}$ \cite{Salton}. When comparing different classification models, we use BOW with binary weights as a baseline but also try extensions like tf- or tf-idf-weigthing and bag of token n-grams. In general, BOW models have a rather high representation length while being rather sparse at the same time (many 0~feature scores).

\paragraph{Part-of-speech (POS) n-grams}
Another vector representation is formed by the frequencies of the 500~most frequent POS bi-, tri and four-grams.\footnote{Using spaCy's POS tagger:\href{https://spacy.io/api/annotation\#pos-tagging}{spacy.io/api/annotation{\#}pos-tagging}.}

\paragraph{Contains JJR}
A Boolean feature capturing the presence of a JJR POS tag (comparative adjective). 

\paragraph{Word Embeddings}
We rely on GloVe~\cite{pennington2014glove} embeddings of size~300 to create a dense, low-dimension vector representation of a sentence.\footnote{Using spaCy's \texttt{en\_core\_web\_lg} model: \href{https://spacy.io/models/en\#section-en_core_web_lg}{spacy.io/models/en{\#}section-en{\_}core{\_}web{\_}lg}.} We average all word vectors of a sentence, representing it by kind of a centroid word---a simple method shown to be effective for several tasks \cite{Wieting:2015aa}.

\paragraph{Sentence Embeddings}
Bags of words and average word embeddings lose sequence information, which intuitively should help for (directed) comparison extraction. Sentence embeddings aim to learn representations for spans of text instead of single words by taking sequence information into account. Several methods like FastSent \cite{hill2016learning} or SkipTought \cite{NIPS2015_5950} have been proposed to create sentence embeddings. We use InferSent  \cite{Conneau:2017aa} that learns sentence embeddings similar to word embeddings. A neural network is trained on the Stanford Natural Language Inference (SNLI) dataset \cite{snli:emnlp2015} containing 570,000~English sentence pairs (each labelled as entailment, contradiction, or neutral). InferSent combines the embeddings~$u$ and~$v$ of the two sentences from a sentence pair into one feature vector (containing the concatenation, the element-wise product, and the element-wise difference of~$u$ and~$v$), that is then fed into a fully connected layer and a softmax layer. We use the pre-trained embeddings in our experiments.\footnote{\href{https://github.com/facebookresearch/InferSent}{github.com/facebookresearch/InferSent}}

\paragraph{Dependency-based Features}
\label{sec:lexnet}
The HypeNet method to detect hypernym relations between words \cite{DBLP:conf/acl/ShwartzGD16} combines distributional and dependency path-based methods to create a vector representation for word pairs. The LexNet generalization of HypeNet encodes tries to capture multiple semantic relationships between two words also using dependency path information \cite{DBLP:journals/corr/ShwartzD16}. Since dependency paths have been one of the major sources for comparison extraction in related work from the biomedical domain (see Section~\ref{sec:relwork}), we also include two LexNet-based features in our experiments.

\subparagraph{LexNet (original)} In the original LexNet paper, an LSTM \cite{hochreiter1997long} is used to create path embeddings out of the string paths. Since the details of the LSTM encoder are not mentioned, we tested different architectures and hyper-parameters and achieved the best results with one LSTM layer with 200~neurons, batch size of~128, RMSprop with learning rate~0.01 and 150~epochs, and max pooling with a pool size of~2. A Keras embedding layer is used to create word embeddings of length~100 for the string path components.

In the original study, paths were restricted to a length of four with the first comparison target having to be reachable from the lowest common head of the two targets by following left edges only, the second one by following right edges. With this LexNet (original) restriction, a path was found for only~1,519 of our 5,759~training sentences. 

\begin{figure*}[t]
\centering
\includegraphics[width=.9\textwidth]{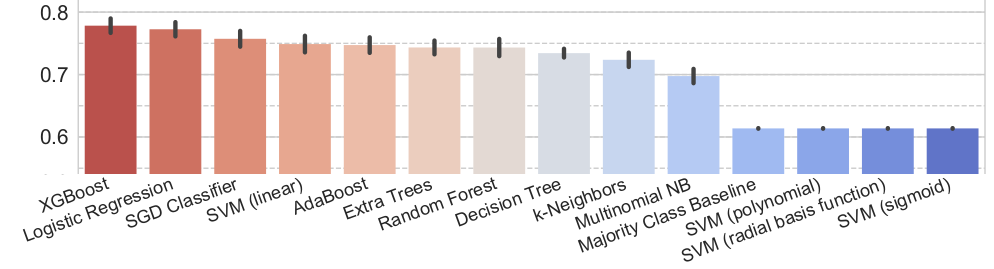}
\caption{Impact of classification models: F1 scores on 5-fold cross validation of various classification algorithms based on a baseline binary bag-of-words representation. The black bars show the standard deviation.}
\label{fig:classification_models}
\end{figure*}

\subparagraph{LexNet (customized)} To overcome the LexNet~(original) coverage issue, we relaxed the restriction by extending the maximal path length to~16 and ignoring edge directions. With this second LexNet (customized) setup, for only 399~training sentences no path was found (assigned to the artificial~NOPATH). 

\section{Experiments}
\label{sec:results}

We conduct classification experiments using several machine learning approaches and representations and analyse the results. We use common performance metrics: precision, recall and F1 per each class and micro-averaged when reporting overall results. 


 

\subsection{Impact of Classification Models}

To identify the best classification algorithm, we used a fixed baseline set of feature representations: a sparse bag-of-words model with binary weights computed on the whole sentence (see Section \ref{sec:features}). We used F1 score to measure the models performance. 

Tree-based methods and linear models worked well. Support Vector Machines with non-linear kernels assigned \class{NONE} to all sentences. As XGBoost and Logistic Regression achieved high F1~scores (see Figure~\ref{fig:classification_models}), no further investigations on the performance of other algorithms were done. A set of hyper-parameters for XGBoost was tested using exhaustive grid search and randomized search but with no significant performance increase. For the futher experiments, we selected XGBoost with 1,000~estimators. The main idea behind boosting is to fit weak learners (i.e., classifiers only performing slightly better than random guessing) sequentially on modified versions of the data subsequently combining them to produce the final prediction. The \textit{XGBoost} boosting method used here is \textit{gradient boosting} \cite{friedman2001greedy} with \textit{decision trees} as learners. In gradient boosting, \textit{$G_{m+1}$} is fitted on the residuals of \textit{$G_m$}. Thus, each following tree tries to improve on the training examples on which the previous learner was weak.

In our experiments, we also tried various neural classification models based on neural network, such as recurrent neural networks, e.g. LSTM~\cite{hochreiter1997long} and simpler feed-forward architectures. However, none of them worked better than the simpler classifiers presented in this paper. We attribute this to the size of our training dataset.

\subsection{Impact of Feature Representations}

\begin{figure*}[htbp]
    \centering
    \includegraphics[width=.90\linewidth]{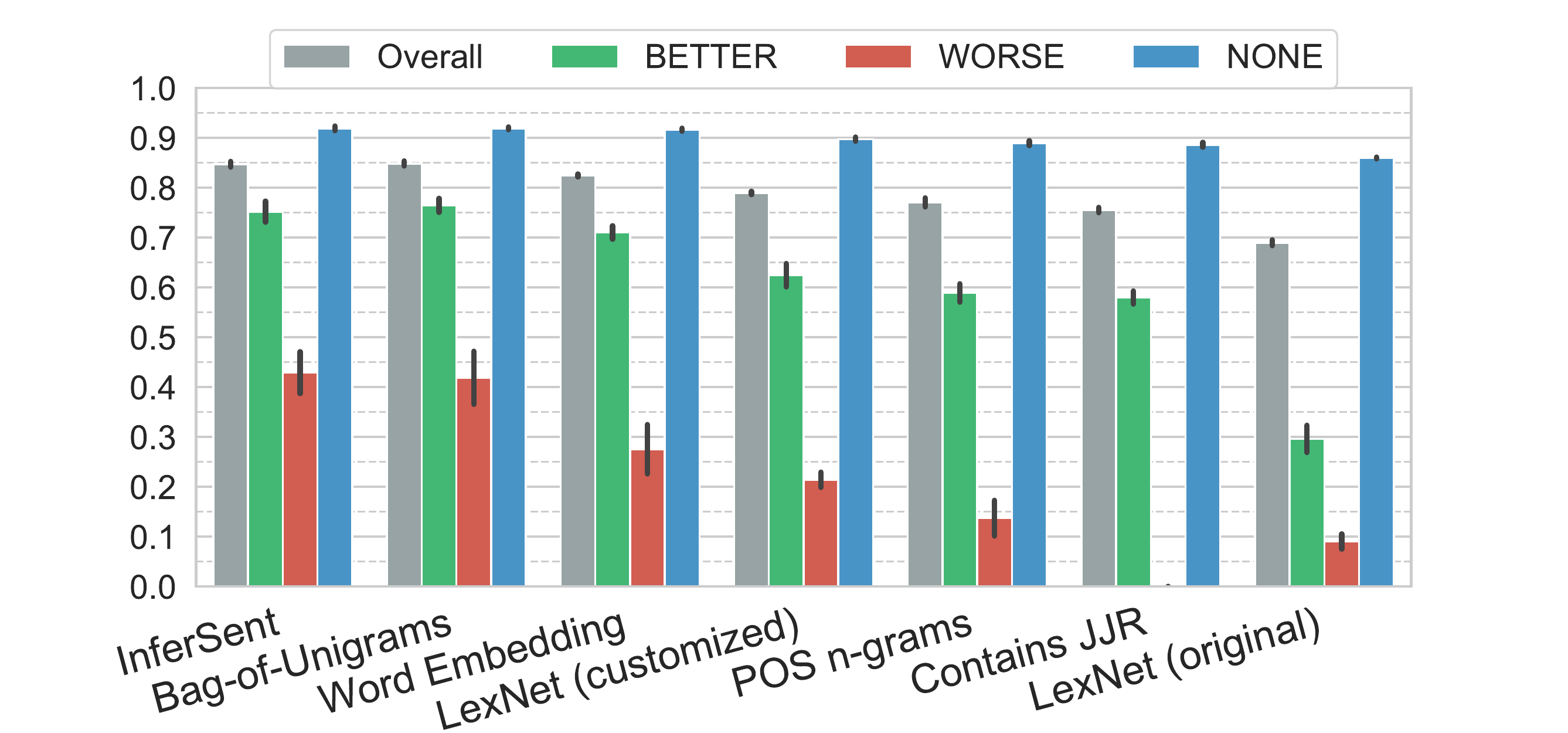}
    \caption{Impact of feature representation: F1 scores of sentence classification model based on XGBoost. The black bars indicate the standard deviation in the 5-fold cross validation.} 
    \label{fig:3_f1}
\end{figure*}

The classification results of the best-performing feature configurations in our three-class scenario are presented in Figure \ref{fig:3_f1}. Each feature was tested and evaluated using five stratified folds. The black bars show the standard deviation. All scores were calculated with scikit-learn's metric module. All features except for the LexNet (original) used the middle part of the sentence and left the objects untouched. In the LexNet features, the comparison targets were replaced with \textit{OBJECT\_A} and \textit{OBJECT\_B}, whereas LexNet (original) used the full sentence.

\begin{table}[b]
\caption{Performance (F1) of the best classifier-based model compared to the rule-based baseline.}
   \scalebox{0.95}{
\small
    \vspace*{1mm}
    \centering
    \begin{tabular}{@{}lcccc@{}}
    	\toprule[0.02cm]
    	\bf Model & {\small \class{BETTER}} & \class{WORSE} & \class{NONE} & \bf{\class{ALL}} \\ \midrule
    	\addlinespace[1ex]	
    	Rule-based Baseline & 0.65 & \bf 0.44 & 0.90 & 0.82 \\
    	InferSent+XGBoost & \bf 0.75 & 0.43 & \bf 0.92 & \bf 0.85 \\
    	\bottomrule
    \end{tabular}
    }
    \label{tab:baseline_classification}
\end{table}

The best single feature (InferSent of the text between objects) yields an overall F1 score 3~points above the baseline with known compared objects positions. The worst single feature (LexNet (original)) scores 12~points below the baseline (see Section~\ref{subsec:rule-based}). Bag-of-Unigrams (F1 score 0.848) and InferSent (F1 score 0.842) deliver roughly equal results.

Despite the fact that only 1,519 sentences got a path embedding for LexNet (original), the feature is able to predict some sentences correctly (F1 score of 0.75 on this subset). This indicates that this feature setup is reasonable and would probably work well if it had a higher coverage.

To our surprise, combining feature representations did not help, i.e., we were not able to exceed over the score of the single best representation (InferSent on the sentence middle part) in any setup, which is why we do not report results on combinations.

Using the full sentence worked second best. Adding the beginning and/or ending part of the sentence did not increase the F1 score at all, no matter if the same or other representation type than the one for the middle part is used. Using the beginning and ending part alone never resulted in an F1 score above the baseline. Similarly, replacing or removing the objects did not increase the score significantly. In most cases, the difference in the F1 score between no replacement/removal and the best replacement/removal strategy was only reflected in the third or fourth decimal place. Hence, the actual objects are not important at all for the classification, which hints at the domain-independence of the dataset. This is also supported by the fact that adding the word vectors of the comparison targets as features did not increase the result in any configuration.

An interesting observation is that the simple bag-of-words model performs equal to or better than the majority of the more complex models in this setup. 
 
\subsection{Comparison to a Rule-based Baseline}\label{subsec:rule-based}

As a rule-based baseline, we adapt the closest classification approach to ours introduced by~\citet{ganapathibhotla-liu2008}. Given a comparative sentence and a pair of the objects being compared, the model decides which one is superior based on the author's opinion. It distinguishes two types of comparatives: opinionated (with explicit preference: \textit{better}, \textit{worse}, etc.) and with context-dependent opinions (implicit preference: \textit{lower}, \textit{higher}, etc.). Classification is performed based on the list of the opinion words considering an opinion orientation borrowed from the work by~\citet{hu:04}. However, our task is different in two aspects. First, we classify  sentences in three not two classes. Second, we identify a comparison direction, i.e., infer a superior object, in a single sentence (and not an overall subjective opinion) without having access to additional context assuming extraction of the objective information. As the authors did not share their code and data, we fetched comparative adjectives and adverbs from open language learning web resources, e.g., sparklebox.co.uk. 
Then we manually organized them in two lists indicating whether the sentence's left-hand located object superior to the right-hand (\textit{better}, \textit{cheaper}, \textit{easier}, etc.) one or not (\textit{worse}, \textit{harder}, \textit{lower}, etc.).  We classify sentences containing a keyword from the first list (74~words in total) as \class{BETTER}, from the second list (63~words) as \class{WORSE} and \class{NONE} with no keywords found. We added negation rules to invert the label if the keyword is preceded by \textit{not} or the second compared object by \textit{but}.

\begin{table}[t]
 \small
       \caption{Cross-domain evaluation in terms of total F1 for all classes (best results per row in bold).}
    \vspace*{1mm}
    \centering
    \begin{tabular}{@{}lp{1.1cm}p{1.1cm}p{1.1cm}@{}}
    	\toprule
    	Train \textbackslash Test & CompSci & Brands & Random \\ \midrule
    	\addlinespace[1ex]	
    	CompSci & 0.82 & \textbf{0.84} & \textbf{0.84} \\ \midrule
    	Brands & 0.76 & \textbf{0.83} & \textbf{0.83} \\ \midrule
    	Random & 0.79 & 0.84 & \textbf{0.86} \\ 
    	\bottomrule
    \end{tabular}
    \label{tab:crossdomain}
\end{table}

\begin{table*}[htb]
\footnotesize
\caption{Examples of XGBoost errors with the InferSent features. Confidence shows the confidence of the annotators and is calculated as (judgments for majority class) / (total judgments).}
\begin{tabularx}{\linewidth}{@{}lXrrr@{}}
\toprule
 & Sentence & Predicted & Gold & Confidence \\ \midrule
1& Is \textbf{Python} better than \textbf{Perl}? & \class{BETTER} & \class{NONE} & 0.6\\ 

2& Is \textbf{Microsoft} better because of \textbf{Apple}? & \class{BETTER} & \class{NONE} & 1.0\\ 
 
3& \textbf{Microsoft} is the devil but \textbf{Sony} truly isn't any better. & \class{WORSE} & \class{NONE} & 1.0\\ 

4& \textbf{Python} is much better suited as a "glue" language, while \textbf{Java} is better characterized as a low-level implementation language. & \class{BETTER} & \class{NONE} & 1.0\\ 
 
5& Its Azure PaaS/IaaS platform hasn't overtaken \textbf{Amazon} yet in market share, but \textbf{Microsoft} has enjoyed nine straight quarters of growth at 10 percent or better & \class{NONE} & \class{WORSE} & 1.0\\ 
 
 6& arrrggghh...\textbf{Python} is a terrible language - only \textbf{Perl} sucks worse. & \class{WORSE} & \class{BETTER} & 1.0\\ 
 
7&  Good to see again a \textbf{Renault} ahead of a \textbf{Ferrari}. & \class{NONE} & \class{BETTER} & 1.0\\ 
\bottomrule

\end{tabularx}

\label{tbl:3_mistakes_se}

\end{table*}

A comparison of the best statistical classifier with this rule-based baseline is presented in Table~\ref{tab:baseline_classification}. The statistical model substantially outperforms the rule-based baseline for the \class{BETTER} and \class{NONE} classes while being comparable for the \class{WORSE} class. The overall improvement of the statistical model over the rule-based approach is about 3 points in terms of F1 score (0.85 as the best achieved performance). Furthermore, note that reported performance of the rule-based model could be a bit inflated as building of the dataset involved the use of similar cue words as those used in this baseline (cf. Section~\ref{sec:dataset}) even though these cue word lists were build independently. 

\subsection{Cross-domain Evaluation}

Table~\ref{tab:crossdomain} presents results of a cross-domain evaluation of our models. As one can observe our model shows remarkably high cross-domain transfer with some out-of-domain combinations outperforming in-domain training, e.g., CompSci-Brands. While a substantial drop is observed for a few other domain pairs, e.g., Random-CompSci, the performance is still well above the majority class baseline suggesting that some knowledge transfer happened even in these cases and comparative argumentation is not highly domain-dependent. 

Similarly, we applied the rule-based baseline to three domains independently and obtained F1 of~0.80 for CompSci, 0.81~for Brands and 0.84~for Random domains.

\subsection{Error Analysis}
\label{sec:error_analysis}

The \class{WORSE} appeared to be the hardest class to recognize: 1,311 sentences were incorrectly classified. We look at comparing the performance of InferSent and LexNet (customized) thoroughly. Both features caused the same errors on 607~sentences. The InferSent feature made 220~additional errors, while the LexNet feature made~484. Surprisingly, the majority of errors was made on sentences with a high annotation confidence: 425~of the shared errors were made on sentences with a confidence of one. InferSent made 156~errors on highly confident sentences, while LexNet made~356. Examples of errors made by the InferSent feature are presented in Table \ref{tbl:3_mistakes_se}. 

The first two sentences look comparative, but they are questions. Despite annotation of questions as \class{NONE} as explicitly stated in the guidelines, InferSent frequently classified questions as comparative. Sentences three and four are comparative, but they have no clear ``winner'' of the comparison. The guidelines instructs that only sentences with obvious ``winners'' should be labeled with \class{BETTER} or \class{WORSE}. InferSent was not able to learn this restriction. Sentence six has three negative words in it. Sentence seven is hard to classify, as it does not contain any comparative cue word.

The LexNet feature made errors in fairly simple sentences like \textit{Right now Apple is worse than Microsoft ever was}. While InferSent's errors can be coarsely grouped, the errors made by LexNet seem to be more random. We assume that the amount of training data for the neural network encoder is not sufficiently large. However, the overall result of LexNet indicates that the encoder trained on more data would likely yield satisfactory results. The performance for LexNet path embeddings shows that this is a reasonable way to encode sentences. The original setup found only paths for 26\%~of the sentences, yet it yielded an F1 score 8 points above the baseline. The customization made it even more powerful. While we expected that a combination of LexNet features and one of the other features like InferSent would be beneficial, as they encode different information (lexical and syntactical), this turned out to be not the case. 

We explain the relatively low performance of all models on the \class{WORSE} class by the fact that people tend to more often refer to use lexical \class{BETTER}-constructions (when the firstly mentioned compared object is favored) than \class{WORSE}-constructions, similarly to many opinion mining datasets, where the positive class is observed more frequently. Besides, the tested models do not use explicit representations of negations, which may lead to a confusion of the \class{BETTER} and \class{WORSE} classes.

\section{Conclusion}
\label{sec:conc} 

We tackle the task of identifying comparative sentences and categorizing the contained preference. Comparisons are a special kind of argumentative premise and can be deployed in constructing pro/con argumentation to support an informed choice. As our contributions, we \Ni create the \corpusname corpus of 7,199~sentences from diverse domains (27\%~of the sentences being comparative and having an annotated preference direction), and \Nii we evaluate several feature-based supervised approaches on our new corpus.

In our experiments, it turned out that the words between two compared items in a sentence are the most important for detecting comparisons and categorizing the preference direction. 

The best classifier has already been integrated in a system that is able to efficiently mine comparative sentences from web-scale sources and to identify the direction of the comparisons: CAM---the comparative argumentative machine~\citet{schildwachter2019answering}. CAM mines sentences from the web-scale Common Crawl and uses them to argumentatively compare objects specified by a user (e.g., whether Python is better than MATLAB for NLP).\footnote{\href{https://ltdemos.informatik.uni-hamburg.de/cam/}{ltdemos.informatik.uni-hamburg.de/cam/}}

Promising directions for future work are exploiting neural classification approaches, integrating features based on contextualized word representations~\cite{peters-etal-2018-deep,devlin2018bert}, and better handling direction shifters like negations and complex implicit syntactic comparative constructions.


\section*{Acknowledgments}

This work has been supported by the Deutsche Forschungsgemeinschaft (DFG) within the project ``Argumentation in Comparative Question Answering (ACQuA)'' (grant BI 1544/7-1 and HA 5851/2-1) that is part of the Priority Program ``Robust Argumentation Machines (RATIO)'' (SPP-1999).

\bibliography{conll2018}
\bibliographystyle{acl_natbib_nourl}

\end{document}